\documentclass[graybox]{svmult}
\usepackage{mathptmx}       % selects Times Roman as basic font
\usepackage{helvet}         % selects Helvetica as sans-serif font
\usepackage{courier}        % selects Courier as typewriter font
\usepackage{type1cm}        % activate if the above 3 fonts are not available on your system
\usepackage{makeidx}         % allows index generation
\usepackage{graphicx}        % standard LaTeX graphics tool when including figure files
\usepackage{epstopdf}
\usepackage{algorithmic}
\usepackage{cite}
\usepackage{multicol}        % used for the two-column index
\usepackage{multirow}
\usepackage[bottom]{footmisc}% places footnotes at page bottom
\usepackage{amsmath}
\usepackage{float}
\usepackage[sort&compress,square,comma,authoryear]{natbib}
\makeindex
\makeatletter
\floatstyle{ruled}
\newfloat{algorithm}{tbp}{loa}
\providecommand{\algorithmname}{Algorithm}
\floatname{algorithm}{\protect\algorithmname}
\makeatother
\begin{document}
	\title*{An Online RFID Localization in the Manufacturing Shopfloor}
	\author{Andri Ashfahani, Mahardhika Pratama, Edwin Lughofer, Qing Cai and Huang Sheng}
	\institute{Andri Ashfahani \at Nanyang Technological University, Singapore, \email{andriash001@e.ntu.edu.sg}
		\and Mahardhika Pratama \at Nanyang Technological University, Singapore, \email{mpratama@ntu.edu.sg}
		\and Edwin Lughofer \at Johannes Kepler University Linz, Austria, \email{edwin.lughofer@jku.at}
		\and Qing Cai \at Nanyang Technological University, Singapore, \email{506183509@QQ.com}
		\and Huang Sheng \at Singapore Institute of Manufacturing Technology, Singapore, \email{shuang@SIMTech.a-star.edu.sg}}
	\authorrunning{(Andri Ashfahani et al.)}
	\maketitle
	\abstract{Radio Frequency Identification technology has gained popularity for cheap and easy deployment. In the realm of manufacturing shopfloor, it can be used to track the location of manufacturing objects to achieve better efficiency. The underlying challenge of localization lies in the non-stationary characteristics of manufacturing shopfloor which calls for an adaptive life-long learning strategy in order to arrive at accurate localization results. This paper presents an evolving model based on a novel evolving intelligent system, namely evolving Type-2 Quantum Fuzzy Neural Network (eT2QFNN), which features an interval type-2 quantum fuzzy set with uncertain jump positions. The quantum fuzzy set possesses a graded membership degree which enables better identification of overlaps between classes. The eT2QFNN works fully in the evolving mode where all parameters including the number of rules are automatically adjusted and generated on the fly. The parameter adjustment scenario relies on decoupled extended Kalman filter method. Our numerical study shows that eT2QFNN is able to deliver comparable accuracy compared to state-of-the-art algorithms.}
	
	\section{Introduction}
	\label{sec:1}
	Radio Frequency Identification (RFID) technology has been used to manage objects location in the manufacturing shopfloor. It is more popular than similar technologies for object localization, such as Wireless Sensor Networks (WSN) and WiFi, due to the affordable price and the easy deployment \citep{ni2011rfid,yang2016rfid}.
	
	In the Maintenance, Repair, and Overhaul (MRO) industry, for example, locating the equipments and trolleys manually over the large manufacturing shopfloor area results in time-consuming activities and increases operator work-load. Embracing RFID technology for localization will help companies improve productivity and efficiency in the industry 4.0. Instead of manually locating the tool-trolleys, RFID localization is utilized to monitor the location real-time. Despite much work and progress in RFID localization technology, it is still challenging problems. The key disadvantage of RFID is that it has low quality signal, which is primarily altered by the complexity and severe noises in the manufacturing shopfloors \citep{chai2017reference}.
	
	Generally, an RFID localization system comprises of three components, i.e. RFID tag, RFID reader, and the data processing subsystem. The reader aims to identify the tag ID and obtain the received signal strength (RSS) information from tags. An object's location can be estimated by observing the RSS. However, the RSS quality in the real-world is very poor, moreover, it keeps changing over time. As an illustration, although RFID tag is used in the static environment, the RSS keeps changing over time. Moreover, a minor change in the surrounding area can greatly fluctuate the RSS. The major factors causing the phenomena are multipath effect and interference. Therefore, obtaining the accurate location relying on RSS information is a hard task. In several works, those challenges are tackled by employing computational techniques, thus the objects location can be achieved accurately \citep{ni2011rfid}.
	
	There are several techniques which can be utilized to estimate the object's location. First of all, the distance from an RFID tag to an RFID reader can be calculated via the two-way radar equation for a monostatic transmitter. It can be obtained easily by solving the equation. Another approach, LANDMARC, is proposed for the indoor RFID localization \citep{ni2004landmarc}. It makes us of $K$ reference tags, and then it evaluates the RSS similarity between reference tags and object tags. A higher weight will be assigned to the reference tags which posses the similar RSS information to the object tags. In the realm of machine learning, support vector regression (SVR) is implemented for the indoor RFID localization \citep{chai2017reference}. It is one-dimensional method and is designed for stationary objects in the small area. Another way to obtain better accuracy is by employing Kalman filter (KF), it has been demonstrated to deal with wavelength ambiguity of the phase measurements \citep{soltani2015enhancing}.
	
	The strategy to estimate the object location via the so-called radar equation is easy to execute. However, the chance to obtain acceptable accuracy is practically impossible due to the severe noises. LANDMARC manage to improve the localization accuracy. Nonetheless, it is difficult to select the reference tag properly in the industrial environments where interference and multipath effect occurred. Improper selection of reference tags can alter the localization accuracy. Similarly, SVR is also designed to address object localization problem in the small area, it even encounters an over-fitting problem. Meanwhile, integrating KF in some works can improve localization accuracy and it also has low computational cost. KF has better robustness and good statistical properties. However, the requirement to calculate the correlation matrix burdens the computation. In addition, the use of KF is also limited by the nonlinearity and non-stationary condition of the real-world \citep{oentaryo2014online,chai2017reference}. The data generated from a non-stationary environment can be regarded as the data stream \citep{pratama2017panfis++}.
	
	Evolving intelligent system (EIS) is an innovation in the field of computational intelligent to deal with data stream \citep{angelov2006evolving,lughofer2008flexfis}. EIS has an open structure, it implies that it can starts the learning processes from scratch or zero rule base. Its rules are automatically formed according to the data stream information. EIS adapts online learning scenario, it conducts the training process in a single-pass mode \citep{pratama2016evolving}. EIS can either adjusts the network parameters or generates fuzzy rule without retraining process. Hence, it is capable to deal with the severe noises and the systems dynamics \citep{pratama2014panfis,lughofer2015generalized,pratama2016evolving}. In several works, the gradient descent (GD) is utilized to adjust the EIS parameters. However, it  is vulnerable to noises due to its sensitivity \citep{oentaryo2014online}. In the premise part, the Gaussian membership function (GMF) is usually employed to capture the input features of EIS. The drawback of GMF is its inadequacy to detect the overlaps between classes. Several works are employed quantum membership function (QMF) to tackle the problem \citep{purushothaman1997quantum,chen2008efficient,lin2006self}. Nevertheless, it is type-1 QMF which is lack of robustness to deal with uncertainties in real-world data streams \citep{pratama2016evolving}.
	
	This research proposes an EIS, namely evolving Type-2 Quantum Fuzzy Neural Network (eT2QFNN). The eT2QFNN adopts online learning mechanism, it processes the incoming data one-by-one and the data is discarded after being learned. Thus, eT2QFNN has high efficiency in terms of computational and memory cost. The eT2QFNN is encompassed by two learning policies, i.e. the rule growing mechanism and parameters adjustment. The first mechanism enables eT2QFNN to start the learning process from zero rule base. It can automatically add the rule on demands. Before a new rule is added to the network, it is evaluated by a proposed formulation, namely modified Generalized Type-2 Datum Significance (mGT2DS). The second mechanism performs parameters adjustment whenever a new rule is not formed. It aims to keep the network adapted to the current data stream. This mechanism is accomplished by decoupled extended Kalman filter (DEKF). It is worth noting that the DEKF algorithm performs localized parameter adjustment, i.e. each rule can be adjusted independently \citep{oentaryo2014online}. In this research, the adjustment is only undertaken on a winning rule, i.e. a rule with the highest contribution. Therefore, DEKF is more efficient than extended Kalman filter (EKF), and yet it still preserves the EKF performance. On the premise part, the interval type-2 QMF (IT2QMF) is proposed to approximate the desired output. It worth noting that it is a universal function approximator which has been demonstrated by several researchers\citep{purushothaman1997quantum,lin2006self,chen2008efficient}. Moreover, it is proficient to form a graded class partition, such that the overlaps between classes can be identified \citep{chen2008efficient}. 
	
	The major contributions of this research are summarized as follows: 1) This research proposes the IT2QMF with uncertain jump position. It is the extended version of QMF. That is, IT2QMF has the interval type-2 capability in terms of incorporating data stream uncertainties. 2) The eT2QFNN is equipped with rule growing mechanism. It can generate its rule automatically in the single-pass learning mode, if a condition is satisfied. The proposed mGT2DS method is employed as the evaluation criterion. 3) The network parameter adjustment relies on DEKF. The mathematical formulation is derived specifically for eT2QFNN architecture. 4) The effectiveness of eT2QFNN has been experimentally validated using real-world RFID localization data.
	
	The remainder of this book chapter is organized as follows. Section II presents the RFID localization system. The proposed type-2 quantum fuzzy membership function and the eT2QFNN network architecture are presented in section III. Section IV presents the learning policies of eT2QFNN and the DEKF for parameter adjustment are discussed. Section V provides empirical studies and comparisons to state-of-the-art algorithms to evaluate the efficacy of eT2QFNN. Finally, section VI concludes this book chapter.
	
	\section{RFID Localization System}
	\label{sec:2}
\begin{figure}[t!]
		\begin{centering}
			\includegraphics[scale=0.35]{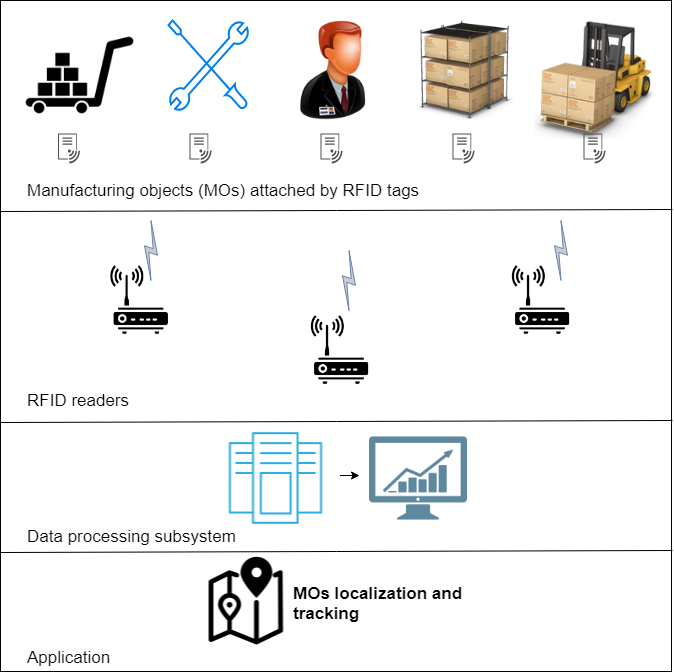}
			\par\end{centering}
		\caption{Architecture of the eT2QFNN}
		\label{fig:rfid}
	\end{figure}

	RFID localization technology has three major components, i.e. RFID tags, RFID readers, and data processing subsystem. There are two types of RFID tags, i.e. the active and passive tag. The active RFID tag is battery-powered and has its own transmitter. It is capable of sending out beacon message, i.e. tag ID and RSS information, actively at specified time window. The transmitted signal can be read up to $300 \thickspace m$ radius. In contrast, passive tag does not have independent  power source. It exploits the reader power signal, thus it cannot actively send the beacon message in a fixed period of time. The signal can only be read around $1 \thickspace m$ radius from the reader, which is very small compared to the manufacturing shopfloor. After that, the transmitted signal is read by the RFID reader. And then it is propagated to the data processing subsystem where the localization algorithm is executed \citep{ni2011rfid}. The configuration of RFID localization is illustrated in the Fig. \ref{fig:rfid}.
	
	The RSS information can be utilized to estimate the object location. It can be achieved by solving the two-way radar equation for a monostatic transmitter as per \eqref{eq:1}. The variables are explained as follows. $P_{T}$, $G_{T}$, $\lambda$, $\sigma$, and $R$ are the reader signal power, the antenna gain, carrier wavelength, tag radar cross-section, and the distance between reader and tag, respectively. However, due to the occurrence of multipath effect and interference, the RSS information becomes unreliable and keeps changing over time. Consequently, the satisfying result cannot be obtained \citep{ni2011rfid,chai2017reference}. Instead of employing \eqref{eq:1} to locate the RFID, this research utilizes eT2QFNN to process the RSS information to obtain the precise object location.
	\begin{equation}\label{eq:1}
	P_{R}=\frac{G_{T}^{2}\lambda^{2}\sigma}{(4\pi)^{2}R^{4}}\thickspace(\mathrm{watts})
	\end{equation}
	
	One of many objectives of this research is to demonstrate the eT2QFNN to deal with the RFID localization problem. The RSS informations of reference tags are utilized to train the network. The reference tags are tags placed at several known and static positions. The eT2QFNN can estimate the new observed tags location according to its RSS. It worth noting that eT2QFNN learning processes are achieved in the evolving mode, it keeps the network parameters and structure adapted to the current data stream. This benefits the network to deal with multipath effect and interference occurred in the manufacturing shopflor.
	
	Now suppose there are $I$ reference tags which are deployed in $M$ locations, then the RSS measurement vector at $n$th time-step can be expressed as $X_{n}=[\begin{array}{ccccc} x_{1} & \ldots & x_{i} & \ldots & x_{I}\end{array}]^T$. Afterward, the network outputs for RFID localization problem can be formulated into a multiclass classification problem. As an illustration, if there exist $M=4$ reference tags deployed in the shopfloor it will indicates that the number of classes is equal to 4.
	
	\section{eT2QFNN Architecture}
	\label{sec:3}
\begin{figure}[t!]
		\begin{centering}
			\includegraphics[scale=0.335]{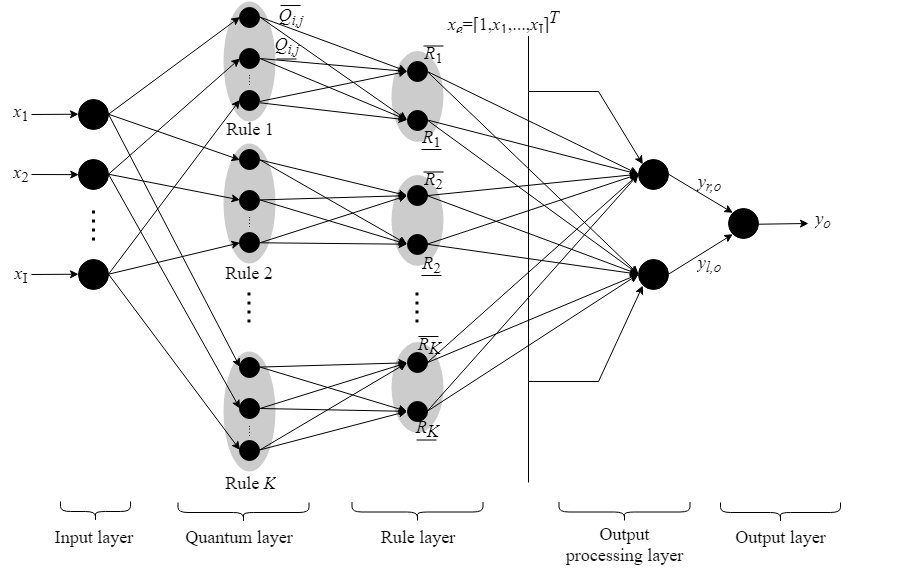}
			\par\end{centering}
		\caption{The RFID localization system}
		\label{fig:et2qfnnfig}
	\end{figure}

	This section presents the network architecture of eT2QFNN. The
	network architecture, as illustrated in the Fig. \ref{fig:et2qfnnfig}, consists of a five-layer,
	multi-input-single-output (MISO) network structure. It is systematized into $I$ input features, $M$ outputs nodes and $K$-term nodes for each input feature. The rule premise
	is compiled of IT2QMF and is expressed as follows:
	\begin{equation}
	\mathrm{\textbf{R}}_{j}:\thickspace\mathrm{\textbf{IF}}\thickspace x_{1}\thickspace\mathrm{\thickspace is\thickspace close\thickspace to}\thickspace\widetilde{Q}_{1j}\thickspace\mathrm{and\ldots and}\thickspace x_{I}\thickspace\mathrm{is\thickspace close\thickspace to}\thickspace\widetilde{Q}_{IK},\thickspace\mathrm{\textbf{THEN}}\thickspace y_{j}^{o}=X_{e}\widetilde{\Omega}_{j}
	\end{equation}
	where $x_{i}$ and $y_{j}^{o}$ are the $i$th input feature and the regression
	output of the $o$th class in the $j$th rule, respectively. $\widetilde{Q}_{ij}=[\overline{Q}_{ij},\underline{Q}_{ij}]$
	denotes the set of upper and lower linguistic term of IT2QMF, $X_{e}$
	is the extended input and $\widetilde{\Omega}_{j}=[\overline{\Omega}_{j},\underline{\Omega}_{j}]$
	is the set of upper and lower consequent weight parameters which are defined as $\widetilde{\Omega}_{j}\in\Re^{M\times2(I+1)}$.
	
\begin{figure}[t!]
		\begin{centering}
			\includegraphics[scale=0.5]{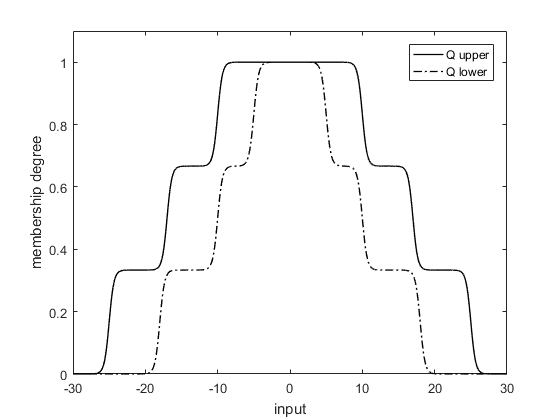}
			\par\end{centering}
		\caption{Interval type-2 quantum membership function with $n_{s}=3$}
		\label{fig:qmf}
	\end{figure}

	The membership function applied in this study is different
	from the typical QMF and GMF. The QMF concept is extended into interval
	type-2 membership function with uncertain jump position. Thus, the
	network can identify overlaps between classes and capable to deal
	with the data stream uncertainties. The IT2QMF output of the $j$th rule
	for the $i$th input feature is given in \eqref{eq:2}
	\begin{eqnarray}\label{eq:2}
	\widetilde{Q}_{ij}(x_{ij},\beta,m_{ij},\widetilde{\theta}_{ij})&=& \frac{1}{n_{s}}\sum_{r=1}^{n_{s}}\left[\left(\frac{1}{1+\exp(-\beta x_{i}-m_{ij}+|\widetilde{\theta}_{ij}^{r}|)}\right)U(x_{i};-\infty,m_{ij})\right. \nonumber\\
	&&+\left.\left(\frac{\exp(-\beta(x_{i}-m_{ij}-|\widetilde{\theta}_{ij}^{r}|))}{1+\exp(-\beta(x_{i}-m_{ij}-|\widetilde{\theta}_{ij}^{r}|))}\right)U(x_{i};m_{ij},\infty)\right]\	
	\end{eqnarray}
	where $m_{ij}$, $\beta$, and $n_{s}$ are mean of $i$th input feature in $j$th rule, slope factor, and number of grades, respectively. $\widetilde{\theta}_{ij}=[\overline{\theta}_{ij},\underline{\theta}_{ij}]$
	is the set of uncertain jump position, it is defined as $\widetilde{\theta}_{ij}\in\Re^{2\times I\times n_{s}\times K}$.
	The upper and lower jump position is expressed as $\overline{\theta}_{ij}=[\begin{array}{ccc}
	\overline{\theta}_{1j}^{1} & \ldots & \overline{\theta}_{1j}^{n_{s}}\end{array};\ldots;\begin{array}{ccc}
	\overline{\theta}_{Ij}^{1} & \ldots & \overline{\theta}_{Ij}^{n_{s}}\end{array}]$ and $\underline{\theta}_{ij}^r=[\begin{array}{ccc}
	\underline{\theta}_{1j}^{1} & \ldots & \underline{\theta}_{1j}^{n_{s}}\end{array};\ldots;\begin{array}{ccc}
	\underline{\theta}_{Ij}^{1} & \ldots & \underline{\theta}_{Ij}^{n_{s}}\end{array}]$. It is defined that  $\overline{\theta}_{ij}^{r}>\underline{\theta}_{ij}^{r}$, thus the execution of \eqref{eq:2} leads to interval type-2 inference scheme which produces a footprint of uncertainties \citep{pratama2016evolving}, it can be clearly seen in the Fig. \ref{fig:qmf}. The eT2QFNN operation in each layer is presented in the following passages.
	\subparagraph{Input layer} This layer performs no computation.
	The data stream is directly propagated to the next layer. The input at $n$th observation is defined by $X_{n}\in\Re^{1\times I}$. And the output of the $i$th node is given as follows:
	\begin{equation}
	u_{i}=x_{i}
	\end{equation}
	\subparagraph{Quantum layer} This layer performs fuzzification step. IT2QMF is utilized to calculate the membership degrees of $X_{n}$ in each rule. The number the existing rule is denoted as $K$. The quantum layer outputs mathematically can be obtained via \eqref{eq:4} and \eqref{eq:5}.
	\begin{eqnarray}
	\overline{Q}_{ij} =\widetilde{Q}_{ij}(x_{i},\beta,m_{ij},\overline{\theta}_{ij}^{r}) \label{eq:4}\\ 
	\underline{Q}_{ij}=\widetilde{Q}_{ij}(x_{i},\beta,m_{ij},\underline{\theta}_{ij}^{r}) \label{eq:5}
	\end{eqnarray}
	\subparagraph{Rule layer} This layer functions to combine the membership degree of $j$th rule denoted as $\widetilde{R}_{j}=[\overline{R}_{j},\underline{R}_{j}]$, which is known as spatial firing strength.
	This can be achieved by employing product T-norm of IT2QMF, as per \eqref{eq:6} and \eqref{eq:7}. The set of upper and lower firing strengths are expressed as $\overline{R}=[\begin{array}{ccc}
	\overline{R}_{1} & \ldots & \overline{R}_{K}\end{array}]$ and $\underline{R}=[\begin{array}{ccc}
	\underline{R}_{1} & \ldots & \underline{R}_{K}\end{array}]$, respectively.
	\begin{eqnarray}
	\overline{R}_{j} & =\prod_{i=1}^{I}\overline{Q}_{ij} \label{eq:6}\\
	\underline{R}_{j} & =\prod_{i=1}^{I}\underline{Q}_{ij} \label{eq:7}
	\end{eqnarray}
	\subparagraph{Output processing layer} The calculation of the two endpoints output, i.e. $y_{l,o}$ and $y_{r,o}$, is conducted here. These variables represent
	the lower and upper crisp output of the $o$th class, respectively.
	The design factor $[q_{l},q_{r}]$ are employed to convert the interval
	type-2 variable to type-1 variable, this is known as the type reduction procedure. This requires less iterative steps compare to the Karnik Mendel (KM) type reduction procedure \citep{pratama2016et2class}. The design factor will govern the proportion of upper and lower IT2QMF and it is defined such that $q_{l}<q_{r}$. The design factor is adjusted using DEKF, thus the proportion of upper and lower outputs $[y_{l},y_{r}]$ keeps adapting to the data streams uncertainties. The lower and upper outputs are given as:
	\begin{eqnarray}
	y_{l,o} & =\frac{(1-q_{l,o})\sum_{j=1}^{K}\underline{R}_j\thickspace\underline{\Omega}_{jo}x_{e}^{T}+q_{l,o}\sum_{j=1}^{K}\overline{R}_j\thickspace\underline{\Omega}_{jo}x_{e}^{T}}{\sum_{j=1}^{K}(\overline{R}_{j}+\underline{R}_{j})}\\
	y_{r,o} & =\frac{(1-q_{r,o})\sum_{j=1}^{K}\underline{R}_j\thickspace\overline{\Omega}_{jo}x_{e}^{T}+q_{r,o}\sum_{j=1}^{K}\overline{R}_j\thickspace\overline{\Omega}_{jo}x_{e}^{T}}{\sum_{j=1}^{K}(\overline{R}_{j}+\underline{R}_{j})}
	\end{eqnarray}
	where $q_l=[q_{l,1},\dots,q_{l,M}]$, $q_r=[q_{r,1},\dots,q_{r,M}]$ are the design factors of all classes, while $\overline{\Omega}_{jo}=[\overline{w}_{ij}^{o},\ldots,\overline{w}_{(I+1)j}^{o}]$,
	and $\underline{\Omega}_{jo}=[\underline{w}_{ij}^{o},\ldots,\underline{w}_{(I+1)j}^{o}]$ express the upper and lower consequent weight parameters of the $j$th
	rule for the $o$th class. In addition, $x_{e}\in\Re^{(I+1)\times1}$
	is the extended input vector. For example, $X_{n}$ has $I$ input
	features $[x_{1},\ldots,x_{I}]$, then the extended input vector is
	$X_{e}=[1,x_{1},\ldots,x_{I}]$. The entry 1 is included to incorporate
	the intercept of the rule consequent and to prevent the untypical gradient \citep{pratama2016evolving}.
	\subparagraph{Output layer} The crisp network output of the $o$th class is the sum of $y_{l,o}$ and $y_{r,o}$ as per \eqref{eq:10}. Furthermore, if
	the network structure of eT2QFNN is utilized to deal with multiclass classification, the multi-model (MM) classifier can be employed to obtain the final classification decision. The MM classifier splits the multiclass classification problem into $K$ binary sub-problems, then $K$ MISO eT2QFNN is built accordingly. The final class decision is the index number $o$ of the highest output, as per \eqref{eq:11}.
	\begin{eqnarray}
	y_{o}&=&y_{l,o}+y_{r,o}\label{eq:10}\\
	y&=&\begin{array}{cc}
	\underset{o=1,\ldots,M}{\arg\max\thickspace}y_{o}\end{array}\label{eq:11}
	\end{eqnarray}
	\section{eT2QFNN Learning Policy}
	The online learning mechanism of eT2QFNN consists of two scenarios, i.e.
	the rule growing and the parameter adjustment which is executed in every iteration. The eT2QFNN starts it learning process with an empty rule base and keeps updating its parameters and network structure as the
	observation data comes in. The proposed learning scenario is presented
	in the Algorithm 1, while subsections 4.1 and 4.2 further explain the learning scenarios.

	\begin{algorithm}
		\caption{Learning policy of eT2QFNN}
		\textbf{Define}: input-output pair $X_{n}=[x_{1},\dots,x_{I}]^{T},\thickspace T_{n}=[t_{1},\dots,t_{M}]^{T}$,
		$n_{s}$, and $\eta$\\
		\textbf{\textbackslash{}\textbackslash{}Phase 1}: \textbf{Rule Growing Mechanism\textbackslash{}\textbackslash{}}\\
		\textbf{If} $K=0$ \textbf{then}\\
		Initiate the first rule via \eqref{newcenter}, \eqref{firsttheta1}, and \eqref{firstdistance1}\\
		\textbf{else}\\
		Approximate the existing IT2QMF via \eqref{approxsigma}\\
		Initiate a hypothetical rule $\textrm{R}_{K+1}$ via \eqref{newcenter}, \eqref{firsttheta}, and \eqref{eq:12}\\
		\textbf{for $j=1$ to} $K+1$\\
		Calculate the statistical contribution \textbf{$E_{j}$ }via \eqref{e}\\
		\textbf{end for}\\
		\textbf{If }$E_{K+1}\ge\rho\sum_{j=1}^{K}E_{j}$ \textbf{then}\\
		$K=K+1$\\
		\textbf{end if}\\
		\textbf{end if}\\
		\textbf{\textbackslash{}\textbackslash{}Phase 2: Parameter Adjustment using DEKF\textbackslash{}\textbackslash{}}\\
		\textbf{If $K(n)=K(n-1)$ then}\\
		Calculate the spatial firing strength via \eqref{eq:6} and \eqref{eq:7}\\
		Determine the winning rule $j_{w}$ via \eqref{eq:12b}\\
		Do DEKF adjustment mechanism on rule $\textrm{\textbf{R}}_{j_{w}}$ via \eqref{eq15b} and \eqref{eq17b}\\
		Update covariance matrix of the winning rule via \eqref{eq16b}\\
		\textbf{else}\\
		Initialize the new rule consequents weight $\widetilde{\Omega}_{K+1}$ and covariance matrix and as \eqref{eq:12} and \eqref{eq:13}\\
		\textbf{for $j=1$ to} $K-1$ \textbf{do}\\
		$P_{j}(n)=P_{j}(n-1)\left(\frac{K^{2}+1}{K^{2}}\right)$\\
		\textbf{end for}\\
		\textbf{end if}
	\end{algorithm}
	
	\subsection{Rule growing mechanism}
	The eT2QFNN is capable of automatically evolving its fuzzy rule on demands using the proposed mGT2DQ method. First of all, it is achieved by forming a hypothetical rule from a newly seen sample. The initialization of hypothetical rule parameters is presented in the sub-subsection \ref{subsubsec:421}. Before it is added to the network, it is required to evaluate its significance. The significance of the $j$th rule is defined as an $L^2-norm$ of $E_{sig}(j)$ weighted by the input density function $p(x)$ as follows \citep{huang2005generalized}:
	\begin{eqnarray}\label{rulesignificance}
	E_{sig}(j)=||\omega_j||\left(\int_{R^I}\exp(-2||X-m_{j}||^2/\sigma_{j}^2) p(X) dX \right)^{1/2}
	\end{eqnarray}
	From the \eqref{rulesignificance}, it is obvious that the input density $p(X)$ greatly contributes to $E_{sig}(j)$. In practical, it is hard to be calculate a priori because the data distribution is unknown. \cite{huang2003recursive} and \cite{huang2004efficient} calculated \eqref{rulesignificance} analytically with the assumption of $p(X)$ being uniformly distributed. However, \cite{zhang2004efficient} demonstrated that it leads to performance degradation for complex $p(X)$. To overcome this problem, \cite{bortman2009growing} proposed Gaussian mixture model (GMM) to approximate the complicated data stream density. The mathematical formulation of GMM is given as:
	\begin{eqnarray}
		\hat{p}(X)&=&\sum_{h=1}^{H}\alpha_{h}\mathcal{N}(X;v_h,\Sigma_{h})\label{PXGMM}\\
		\mathcal{N}(X;v_{h},\Sigma_{h})&=&\exp(-(X-v_{h})^T\Sigma^{-1}_h(X-v_{h}))\label{gaussianfunction}
	\end{eqnarray}
	where $\mathcal{N}(X;v_h,\Sigma_{H})$ is the Gaussian function of variable $X$ as per \eqref{gaussianfunction}, with the mean vector $v_{h}\in\Re^I$, variance matrix $\Sigma_{h}\in^{I\times I}$, $H$ denotes the number of mixture model, and $\alpha_{h}$ represent the mixing coefficients ($\sum_{h=1}^{H}\alpha_{h}=1; \alpha_{h}>0\forall\thickspace h$).
	
	In the next step, the estimated significance of $j$th rule $\hat{E}_{sig}(j)$ is calculated. \cite{vukovic2013growing} derived the mathematical formulation to obtain $\hat{E}_{sig}(j)$ by substituting \eqref{PXGMM} to \eqref{rulesignificance} and then solving the closed form analytical solution, it yields to the following result:
	\begin{eqnarray}\label{rulesignificanceestimation}
	\hat{E}_{sig}(j)=||\omega_j||(\pi^{I/2}\det(\Sigma_{j})^{1/2}N_jA^T)^{1/2}
	\end{eqnarray}
	where $A=[\alpha_{1},\dots,\alpha_{H}]$ is the vector of GMM mixing coefficients, $\Sigma_{j}$ denotes as the positive definite weighting matrix which is expressed as $\Sigma_{j}=\textrm{diag}(\sigma_{1,j}^2,\dots,\sigma_{I,j}^2)$, and $N_j$ is given as:
	\begin{eqnarray}
	N_j&=&[\mathcal{N}(m_j-v_1;0,\Sigma_j/2+\Sigma_{1}),\thickspace \mathcal{N}(m_j-v_2;0,\Sigma_j/2+\Sigma_{2}),\dots \nonumber \\
	&&\dots,\mathcal{N}(m_j-v_H;0,\Sigma_j/2+\Sigma_{H})]
	\end{eqnarray}
	where $m_j$ is the mean vector of $j$th rule defined as $m_j=[m_{1,j},\dots,m_{I,j}]$. And the GMM parameters $v_h$, $\Sigma_{h}$, and $A$ can be calculated by exploiting $N_{history}$ pre-recorded data. This technique is feasible and easy to implement because the pre-recorded input data is most likely to be stored especially in the era of data stream. The number of pre-recorded data is somewhat smaller than the training data, it is denoted as $N_{history} \ll N$. It is not problem-specific and it can be set to a fixed value \citep{pratama2017panfis++}. In this research, it is set as 50 for simplicity.
	
\begin{figure}[t!]
		\begin{centering}
			\includegraphics[scale=0.50]{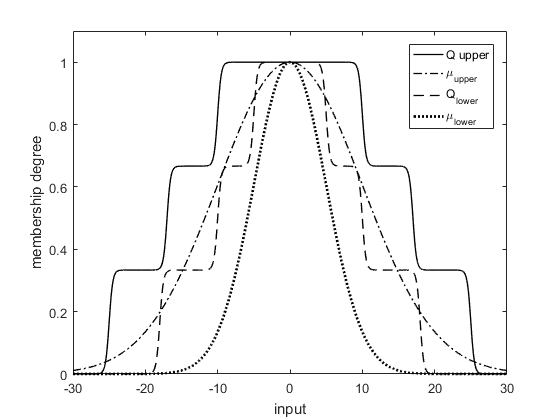}
			\par\end{centering}
		\caption{The comparison result of IT2QMF and IT2GMF}
		\label{fig:quppermuupper}
	\end{figure}

	The method \eqref{rulesignificanceestimation} could not, however, be applied directly to estimate the eT2QFNN rule significance, because eT2QFNN utilizes IT2QMF instead of Gaussian membership function (GMF). The key idea to overcome this problem is by approximating IT2QMF using interval type-2 Gaussian Membership Function (IT2GMF). The mathematical formulation of this approach can be written as follows: 
	\begin{eqnarray}
	\widetilde{Q}_{i,j}(x_{i},\beta,m_{i,j},\widetilde{\theta}_{ij})&\approx&\widetilde{\mu}_{i,j}=\exp\left(-\frac{(x_{i}-m_{i,j})^2}{\widetilde{\sigma}_{i,j}}\right)\label{approxsigma}\\
	\widetilde{\sigma}_{i,j}&=&[\begin{array}{cc}\underline{\sigma}_{i,j}, & \overline{\sigma}_{i,j}\end{array}]\nonumber\\
	\underline{\sigma}_{i,j}=\min\thickspace\underline{\theta}_{i,j};&\thickspace& \overline{\sigma}_{i,j}=\min\thickspace\overline{\theta}_{i,j}\label{sigmaminimum}
	\end{eqnarray}
	The mean of IT2GMF is defined to equal the mean of IT2QMF, i.e. $m_{ij}$. And the width of upper and lower IT2GMF are obtained by taking the minimum value of $\widetilde{\theta}_{ij}$ as per \eqref{sigmaminimum}. By selecting these criteria, the whole area of IT2GMF will be located inside the area of IT2QMF. As illustrated in the Fig. \ref{fig:quppermuupper}, both upper and lower area of IT2GMF covers the appropriate area of IT2QMF. Therefore, this approach can provide a good approximation of IT2QMF. 
	
	Afterward, the proposed method to estimate eT2QFNN rule significance can be derived by executing \eqref{rulesignificanceestimation} with the design factor as per \eqref{e}. In \eqref{EL} and \eqref{ER}, $\overline\Omega_{j}=[\overline\Omega_{j,1},\dots,\overline\Omega_{j,M}]^T$ and $\underline\Omega_{j}=[\underline\Omega_{j,1},\dots,\underline\Omega_{j,M}]^T$ are denoted as the upper and lower consequent weight parameters of all classes, respectively. The variance matrices $\overline{\Sigma}_j$ and $\underline{\Sigma}_j$ are formed of $\overline{\sigma}_{i,j}$ and $\underline{\sigma}_{i,j}$ as per \eqref{varianceesig}, while $\overline{N}_j$ and $\underline{N}_j$ are given in \eqref{Nesig}. The hypothetical rule will be added to the network as a new rule $\mathrm{\textbf{R}}_{K+1}$ if it posses statistical contribution over existing rules. The mathematical formulation of rule growing criterion is given in \eqref{statcont}, where the constant $\rho \in (0,1]$ is defined as the vigilance parameter and in this research it is fixed at 0.65 for simplicity.
	\begin{eqnarray}
	\hat{E}_{j} & = & |\hat{E}_{j,l}|+|\hat{E}_{j,r}|\label{e}\\
	\hat{E}_{j,l} & = & ||q_{l}||\cdot||\overline{\Omega}_{j}||\cdot(\pi^{I/2}\det(\overline{\Sigma}_{j})^{1/2}\overline{N}_jA^T)^{1/2}\nonumber \\
	&& +  (1-||q_{l}||)\cdot||\underline{\Omega}_{j}||\cdot(\pi^{I/2}\det(\underline{\Sigma}_{j})^{1/2}\underline{N}_jA^T)^{1/2}\label{EL}\\
	\hat{E}_{j,r} & = & ||q_{r}||\cdot||\overline{\Omega}_{j}||\cdot(\pi^{I/2}\det(\overline{\Sigma}_{j})^{1/2}\overline{N}_jA^T)^{1/2}\nonumber\\
	&& +  (1-||q_{r}||)\cdot||\underline{\Omega}_{j}||\cdot(\pi^{I/2}\det(\underline{\Sigma}_{j})^{1/2}\underline{N}_jA^T)^{1/2}\label{ER}\\
	E_{K+1}&\geq&\rho\sum_{j=1}^{K} E_{j} \label{statcont}\\
	\overline{\Sigma}_{j}&=&\textrm{diag}(\overline{\sigma}_{1,j}^2,\dots,\overline{\sigma}_{I,j}^2),\thickspace \underline{\Sigma}_{j}=\textrm{diag}(\underline{\sigma}_{1,j}^2,\dots,\underline{\sigma}_{I,j}^2) \label{varianceesig}\\
	\overline{N}_j&=&[\mathcal{N}(m_j-v_1;0,\overline{\Sigma}_j/2+\Sigma_{1}),\thickspace \mathcal{N}(m_j-v_2;0,\overline{\Sigma}_j/2+\Sigma_{2}),\dots \nonumber \\
	&&\dots,\mathcal{N}(m_j-v_H;0,\overline{\Sigma}_j/2+\Sigma_{H})],\nonumber\\
	\underline{N}_j&=&[\mathcal{N}(m_j-v_1;0,\underline{\Sigma}_j/2+\Sigma_{1}),\thickspace \mathcal{N}(m_j-v_2;0,\underline{\Sigma}_j/2+\Sigma_{2}),\dots \nonumber \\
	&&\dots,\mathcal{N}(m_j-v_H;0,\underline{\Sigma}_j/2+\Sigma_{H})] \label{Nesig}
	\end{eqnarray}
	
	\subsection{Parameter adjustment}
	This phase comprises of two alternative strategies. The first strategy is carried out to form a hypothetical rule. It will be added into the network structure if the condition in \eqref{statcont} is satisfied. This strategy is called the fuzzy rule initialization. And the second mechanism is executed whenever \eqref{statcont} is not satisfied. It is aimed to adjust the network parameters according to the current data stream. This is called the winning rule update. These strategies are elaborated in the following sub-subsections.
	
	\subsubsection{Fuzzy rule initialization}
	\label{subsubsec:421}
	The rule growing mechanism first of all is conducted by forming a hypothetical rule according to the current data stream. The data at $n$th time-step $X_n$ is assigned as the new mean of IT2QMF, as per \eqref{newcenter}. And then, the new jump position is achieved via distance-based formulation inspired by \citep{lin2006self}, as per \eqref{firsttheta}. In this research, however, it is modified such that the new distance $\widetilde{\sigma}_{i,K+1}$ is obtained utilizing the mixed mean of GMM $\hat{v}$, as per \eqref{firstdistance}. Thanks to the GMM features which is able to approximate the mean and variance of very complex input. For this reason, instead of using $\widetilde{\sigma}_{i,K+1}$ to calculate the jump position of the first rule in \eqref{firsttheta1}, the eT2QFNN utilizes the diagonal entries of the mixed variance matrix, as per \eqref{firstdistance1}. The constant $\delta_{1}$ is introduced to create the footprint of uncertainty. In this study, it is set $\delta_1=0.7$ for simplicity.
	
	In the next stage, the consequent weight parameters of hypothetical rule are determined. It is equal to the consequent weight of the winning rule as per \eqref{eq:12}. The key idea behind this strategy is to acquire the knowledge of winning rule in terms of representing the current data stream \citep{oentaryo2014online}. The way to select the winning rule is presented in sub-subsection \eqref{subsubsec:422}. Finally, if the hypothetical rule passes the evaluation criterion in \eqref{statcont}, it is added as the new rule ($\mathrm{\textbf{R}}_{K+1}$) and its covariance matrix is initialized via \eqref{eq:13}.
	
	In contrast, a consideration is required to adjust the covariance matrices of other rules, because the new rule formation corrupts those matrices. This phenomenon has been investigated in SPLAFIS \citep{oentaryo2014online}, the research revealed that those matrices need to be readjusted. The proper readjustment technique is achieved by multiplication of those matrices and $\left(\frac{K^{2}+1}{K^{2}}\right)$ as per \eqref{eq:14}. This strategy is signified to take into account the contribution that a new rule would have if it existed from the first iteration. It, therefore, will decrease the corruption effect \citep{oentaryo2014online}.	
	\begin{eqnarray}
	m_{K+1}&=&X_{n} \label{newcenter} \\
	\overline{\theta}_{i,K+1}^{r}&=&\frac{1}{((n_{s}+1)/2)}\cdot r\cdot\overline{\sigma}_{i,K+1}, \nonumber\\
	\underline{\theta}_{i,K+1}^r&=&\frac{1}{((n_{s}+1)/2)}\cdot r\cdot\underline{\sigma}_{i,K+1} \label{firsttheta}\\
	\overline{\sigma}_{i,K+1}&=&|X_{n}-\hat{v}|, \thickspace \underline{\sigma}_{i,K+1}=\delta_1 \cdot \overline{\sigma}_{i,K+1}\label{firstdistance}\\
	\hat{v}&=&\sum_{h=1}^{H}\alpha_{h}\cdot v_{h}\nonumber
	\end{eqnarray}
	\begin{eqnarray}
	\overline{\theta}_{i,1}^{r}&=&\frac{1}{((n_{s}+1)/2)}\cdot r\cdot\overline{\sigma}_{i,1}, \nonumber\\
	\underline{\theta}_{i,1}^r&=&\frac{1}{((n_{s}+1)/2)}\cdot r\cdot\underline{\sigma}_{i,1} \label{firsttheta1}\\
	\overline{\sigma}_{i,1}&=&\hat{\sigma}_{i}, \thickspace \underline{\sigma}_{i,1}=\delta_1 \cdot \overline{\sigma}_{i,1}\label{firstdistance1}\\
	\hat{\Sigma}&=&\sum_{h=1}^{H}\Sigma_{h}\cdot v_{h}, \thickspace \hat{\Sigma} = \textrm{diag}(\hat{\sigma}_{1}^2,\dots,\hat{\sigma}_{I}^2) \nonumber
	\end{eqnarray}
	\begin{eqnarray}
	\widetilde{\Omega}_{K+1}&=&\widetilde{\Omega}_{j_w}\label{eq:12}\\
	P_{K+1}(n)&=&I_{Z\times Z}\label{eq:13}\\
	P_{j}(n)&=&\left(\frac{K^{2}+1}{K^{2}}\right)P_{j}(n-1)\label{eq:14}
	\end{eqnarray}
	\subsubsection{Winning rule update}
	\label{subsubsec:422}
	The hypothetical rule would not be added to the network structure if it failed the evaluation in \eqref{statcont}. To maintain the eT2QFNN performance, the network parameters is required to be adjusted according to the information provided by the current data stream. In this research, the adjustment is only undertaken on the winning rule which is defined as a rule having the highest average of the spatial firing strength. The mathematical formulation is given in \eqref{eq:12b}. It worth noting that spatial firing strength represents the degree to which the rule antecedent part is satisfied. The rule having higher firing strength possesses higher correlation to the current data stream \citep{pratama2014panfis}, therefore it deserves to be adjusted.
	\begin{eqnarray}
	j_{w}&=&\underset{j}{\arg\max\thickspace}\widetilde {R_{j}}\label{eq:12b}\\
	\widetilde{R_{j}}&=&\frac{\overline{R}_{j}+\underline{R}_{j}}{2}
	\end{eqnarray}
	
	Previously, DEKF is employed to adjust the winning rule parameters of type-1 fuzzy neural network. It is capable of maintaining local learning property of EIS because it can adjust parameters locally \citep{oentaryo2014online}. In this research, DEKF is utilized to update the winning rule parameters of eT2QFNN. The local parameters are classified by rule, i.e. the parameters in the same rule are grouped together. This leads to the formation of block-diagonal covariance matrix $\widetilde{P}(n)$ as per \eqref{14b}. There is only one block covariance matrix updated in each time-step, i.e. $P_{j_w}(n)$. The localized adjustment property of DEKF enhances the algorithm efficiency in terms of computational complexity and memory requirements, moreover it still maintains the same robustness as the EKF \citep{puskorius1994neurocontrol}.
	\begin{equation}\label{14b}
	\widetilde{P}(n)=\left[\begin{array}{ccccc}
	P_{1}(n) & \ldots & 0 & \ldots & 0\\
	\vdots & \ddots &  &  & \vdots\\
	0 &  & P_{j}(n) &  & 0\\
	\vdots &  &  & \ddots & \vdots\\
	0 & \ldots & 0 & \ldots & P_{K}(n)
	\end{array}\right]
	\end{equation}
	
	The mathematical formulations of DEKF algorithm are given in \eqref{eq15b}-\eqref{eq17b}. The designation of each parameter in the equation is as follows. $G_{j_{w}}(n)$ and $P_{j_{w}}(n)$ are the Kalman gain matrix and covariance matrix, respectively. The covariance matrix represents the interaction between each pair of the parameters in the network. $\overrightarrow{\theta}_{j_{w}}(n)$ is the parameter vector of $\mathrm{\textbf{R}}_{j_w}$ at $n$th iteration, it consists of all the network parameters which is about to be adjusted. It is expressed as $\overrightarrow{\theta}_{j_{w}}(n)=[\underline{\Omega}_{j_w}^T, \thickspace \overline{\Omega}_{j_w}^T, \thickspace q_{l}^T, \thickspace q_{r}^T, \thickspace m_{j_w}^T, \thickspace \underline{\theta}_{j_w}^T, \thickspace \overline{\theta}_{j_w}^T]^T$, which is respectively given in \eqref{omegalow}-\eqref{thetaup}. The length of $\overrightarrow{\theta}_{j_{w}}(n)$ is equal to $Z=2\times M\times(2+I)+I\times(2 \times n_s +1)$. The Jacobian matrix $H_{k_{w}}(n)$, presented in \eqref{jacobian}, contains the output gradient with respect to the network parameters, and it is arranged into $Z$-by-$M$ matrix. The gradient vectors are specified in \eqref{gradient} and is calculated using \eqref{ymyw}-\eqref{ymytheta}. The output and target vectors are defined as $y(n)=[\begin{array}{ccc}
	y_{1}(n) & \ldots & y_{M}(n)\end{array}]$ and $t(n)=[\begin{array}{ccc}
	t_{1}(n) & \ldots & t_{M}(n)\end{array}]$. It is utilized to calculate the error vector in \eqref{eq17b}. The last parameters, $\eta$, is a learning rate parameter \citep{puskorius1994neurocontrol}. This completes the second strategy to maintain the network adapted to the current data stream.
	\begin{eqnarray}
	G_{j_{w}}(n)&=&P_{j_{w}}(n-1)H_{j_{w}}(n)[\eta I_{M\times M}+H_{k_{w}}^{T}(n)P_{j_{w}}(n-1)H_{j_{w}}(n)]^{-1}\label{eq15b}\\
	P_{j_{w}}(n)&=&[I_{Z\times Z}-G_{j_{w}}(n)H_{j_{w}}^{T}(n)]P_{j_{w}}(n-1)\label{eq:16}\label{eq16b}\\
	\overrightarrow{\theta}_{j_{w}}(n)&=&\overrightarrow{\theta}_{j_{w}}(n-1)+G_{j_{w}}(n)[t(n)-y(n)]\label{eq17b}
	\end{eqnarray}
	\begin{eqnarray}
	\underline{\Omega}_{j_w}&=&[\underline{\Omega}_{1,j_w}^1,\dots,\underline{\Omega}_{I+1,j_w}^1,\dots,\underline{\Omega}_{1,j_w}^1,\dots,\underline{\Omega}_{I+1,j_w}^M]^T\label{omegalow}\\
	\overline{\Omega}_{j_w}&=&[\overline{\Omega}_{1,j_w}^1,\dots,\overline{\Omega}_{I+1,j_w}^1,\dots,\overline{\Omega}_{1,j_w}^M,\dots,\overline{\Omega}_{I+1,j_w}^M]^T\\
	q_{l}&=&[q_{l,1},\dots,q_{l,M}]^T\\
	q_{r}&=&[q_{r,1},\dots,q_{r,M}]^T\\
	m_{j_w}&=&[m_{1,j_w},\dots,m_{I,j_w}]^T\\
	\underline{\theta}_{j_w}&=&[\underline{\theta}_{1,j_w}^1,\dots,\underline{\theta}_{I,j_w}^1,\dots,\underline{\theta}_{1,j_w}^{n_s},\dots,\underline{\theta}_{I,j_w}^{n_s}]^T\\
	\overline{\theta}_{j_w}&=&[\overline{\theta}_{1,j_w}^1,\dots,\overline{\theta}_{I,j_w}^1,\dots,\overline{\theta}_{1,j_w}^{n_s},\dots,\overline{\theta}_{I,j_w}^{n_s}]^T\label{thetaup}
	\end{eqnarray}
	\begin{eqnarray}
	H_{j_{w}}(n)=\left[\begin{array}{ccccc}
	\frac{\partial y_{1}}{\partial\underline{\Omega}_{j_{w},1}} & \dots & 0 & \dots & 0\\
	0 & \dots & \frac{\partial y_{o}}{\partial\underline{\Omega}_{j_{w},o}} & \dots & 0\\
	0 & \dots & 0 & \dots & \frac{\partial y_{M}}{\partial\underline{\Omega}_{j_{w},M}}\\
	\frac{\partial y_{1}}{\partial\overline{\Omega}_{j_{w},1}} & \dots & 0 & \dots & 0\\
	0 & \dots & \frac{\partial y_{o}}{\partial\overline{\Omega}_{j_{w},o}} & \dots & 0\\
	0 & \dots & 0 & \dots & \frac{\partial y_{M}}{\partial\overline{\Omega}_{j_{w},M}}\\
	\frac{\partial y_{1}}{\partial q_{l,1}} & \dots & 0 & \dots & 0\\
	0 & \dots & \frac{\partial y_{o}}{\partial q_{l,o}} & \dots & 0\\
	0 & \dots & 0 & \dots & \frac{\partial y_{M}}{\partial q_{l,M}}\\
	\frac{\partial y_{1}}{\partial q_{r,1}} & \dots & 0 & \dots & 0\\
	0 & \dots & \frac{\partial y_{o}}{\partial q_{r,o}} & \dots & 0\\
	0 & \dots & 0 & \dots & \frac{\partial y_{M}}{\partial q_{r,M}}\\
	\frac{\partial y_{1}}{\partial m_{j_{w},1}} & \dots & \frac{\partial y_{o}}{\partial m_{j_{w},o}} & \dots & \frac{\partial y_{M}}{\partial m_{j_{w},M}}\\
	\frac{\partial y_{1}}{\partial\underline{\theta}_{j_{w},1}} & \dots & \frac{\partial y_{o}}{\partial\underline{\theta}_{j_{w},o}} & \dots & \frac{\partial y_{M}}{\partial\underline{\theta}_{j_{w},M}}\\
	\frac{\partial y_{1}}{\partial\overline{\theta}_{j_{w},1}} & \dots & \frac{\partial y_{o}}{\partial\overline{\theta}_{j_{w},o}} & \dots & \frac{\partial y_{M}}{\partial\overline{\theta}_{j_{w},M}}
	\end{array}\right]\label{jacobian}
	\end{eqnarray}
	\begin{eqnarray}
	\frac{\partial y_{o}}{\partial\underline{\Omega}_{j_{w},o}}&=&\left[\frac{\partial y_{o}}{\partial\underline{w}_{1,j_{w}}^o},\dots,\frac{\partial y_{o}}{\partial\underline{w}_{I+1,j_{w}}^o}\right]^T, \thickspace \frac{\partial y_{o}}{\partial\overline{\Omega}_{j_{w},o}}=\left[\frac{\partial y_{o}}{\partial\overline{w}_{1,j_{w}}^o},\dots,\frac{\partial y_{o}}{\partial\overline{w}_{I+1,j_{w}}^o}\right]^T,\nonumber\\
	\frac{\partial y_{o}}{\partial m_{j_{w},o}}&=&\left[\frac{\partial y_{m}}{\partial m_{1,j_{w}}^o},\dots,\frac{\partial y_{o}}{\partial m_{I,j_{w}}^o}\right]^T,\nonumber\\
	\frac{\partial y_{o}}{\partial\underline{\theta}_{j_{w},o}}&=&\left[\frac{\partial y_{o}}{\partial\underline{\theta}_{1,j_{w},o}^1},\dots,\frac{\partial y_{o}}{\partial\underline{\theta}_{I,j_{w},o}^1},\dots,\frac{\partial y_{o}}{\partial\underline{\theta}_{1,j_{w},o}^{n_s}},\dots,\frac{\partial y_{o}}{\partial\underline{\theta}_{I,j_{w},o}^{n_s}}\right]^T,\nonumber\\
	\frac{\partial y_{o}}{\partial\overline{\theta}_{j_{w},o}}&=&\left[\frac{\partial y_{o}}{\partial\overline{\theta}_{1,j_{w},o}^1},\dots,\frac{\partial y_{o}}{\partial\overline{\theta}_{I,j_{w},o}^1},\dots,\frac{\partial y_{o}}{\partial\overline{\theta}_{1,j_{w},o}^{n_s}},\dots,\frac{\partial y_{o}}{\partial\overline{\theta}_{I,j_{w},o}^{n_s}}\right]^T\label{gradient}
	\end{eqnarray}
	\begin{eqnarray}
	\frac{\partial y_{o}}{\partial\underline{w}_{1,j_{w}}^o}&=&\left[\frac{(1-q_{l,o})\underline{R}_{j_w}+q_{l,o}\overline{R}_{j_w}}{\sum_{j=1}^K (\underline{R}_{j_w}+\overline{R}_{j_w})}\right]x_{e,i}(n),\nonumber\\ \frac{y_{o}}{\partial\overline{w}_{1,j_{w}}^o}&=&\left[\frac{(1-q_{r,o})\underline{R}_{j_{w}}+q_{r,o}\overline{R}_{j_{w}}}{\sum_{j=1}^{K}(\underline{R}_{j_{w}}+\overline{R}_{j_{w}})}\right]x_{e,i}(n)\label{ymyw}
	\end{eqnarray}
	\begin{eqnarray}
	\frac{\partial y_o}{\partial q_{l,o}}&=&\left[\frac{-\sum_{j=1}^{K}\underline{R}_{j}\underline{\Omega}_{o}+\sum_{j=1}^{K}\overline{R}_{j}\underline{\Omega}_{o}}{\sum_{j=1}^K (\underline{R}_{j_w}+\overline{R}_{j_w})}\right]X_e(n),\nonumber\\ 
	\frac{\partial y_{o}}{\partial q_{r,o}}&=&\left[\frac{-\sum_{j=1}^{K}\underline{R}\overline{\Omega}_{o}+\sum_{j=1}^{K}\overline{R}\overline{\Omega}_{o}}{\sum_{j=1}^{K}(\underline{R}_{j_{w}}+\overline{R}_{j_{w}})}\right]X_{e}(n)
	\end{eqnarray}
	\begin{eqnarray}
	\frac{\partial y_o}{\partial m_{i,j_w}}&=&\frac{\partial y_o}{\partial y_{l,o}} \left[\frac{\partial y_{l,o}}{\partial\underline{R}_{j_w}}\frac{\partial \underline{R}_{j_w}}{\partial m_{i,j_w}}+\frac{\partial y_{l,o}}{\partial\overline{R}_{j_w}}\frac{\partial \overline{R}_{j_w}}{\partial m_{i,j_w}}\right]\nonumber\\
	&+&\frac{\partial y_o}{\partial y_{r,o}} \left[\frac{\partial y_{r,o}}{\partial\underline{R}_{j_w}}\frac{\partial \underline{R}_{j_w}}{\partial m_{i,j_w}}+\frac{\partial y_{r,o}}{\partial\overline{R}_{j_w}}\frac{\partial \overline{R}_{j_w}}{\partial m_{i,j_w}}\right]\\
	\frac{\partial y_o}{\partial \underline{\theta}_{i,j_w}^r}&=&\frac{\partial y_o}{\partial y_{l,o}}\frac{\partial y_{l,o}}{\partial\underline{R}_{j_w}}\frac{\partial \underline{R}_{j_w}}{\partial\underline{\theta}_{i,j_w}^r}+\frac{\partial y_o}{\partial y_{r,o}}\frac{\partial y_{r,o}}{\partial\underline{R}_{j_w}}\frac{\partial \underline{R}_{j_w}}{\partial\underline{\theta}_{i,j_w}^r},\nonumber\\
	\frac{\partial y_o}{\partial \overline{\theta}_{i,j_w}^r}&=&\frac{\partial y_o}{\partial y_{l,o}}\frac{\partial y_{l,o}}{\partial\overline{R}_{j_w}}\frac{\partial \overline{R}_{j_w}}{\partial\overline{\theta}_{i,j_w}^r}+\frac{\partial y_o}{\partial y_{r,o}}\frac{\partial y_{r,o}}{\partial\overline{R}_{j_w}}\frac{\partial \overline{R}_{j_w}}{\partial\overline{\theta}_{i,j_w}^r}\label{ymytheta}
	\end{eqnarray}
	\begin{eqnarray}
	\frac{\partial y_o}{\partial y_{l,o}}&=&\frac{\partial y_o}{\partial y_{l,o}}=1\nonumber\\
	\frac{\partial y_{l,o}}{\partial\underline{R}_{j_{w}}}&=&\frac{(1-q_{l,o})\underline{\Omega}_{j_{w},o}X_{e}(n)}{\sum_{j=1}^{K}(\underline{R}_{j}+\overline{R}_{j})}-\frac{(1-q_{l,o})\underline{R}\underline{\Omega}_{m}X_{e}(n)+q_{l,o}\overline{R}\underline{\Omega}_{m}X_{e}(n)}{(\sum_{j=1}^{K}(\underline{R}_{j}+\overline{R}_{j}))^{2}}\nonumber\\
	\frac{\partial y_{l,o}}{\partial\overline{R}_{j_{w}}}&=&\frac{q_{l,o}\underline{\Omega}_{j_{w},o}X_{e}(n)}{\sum_{j=1}^{K}(\underline{R}_{j}+\overline{R}_{j})}-\frac{(1-q_{l,o})\underline{R}\underline{\Omega}_{m}X_{e}(n)+q_{l,o}\overline{R}\underline{\Omega}_{m}X_{e}(n)}{(\sum_{j=1}^{K}(\underline{R}_{j}+\overline{R}_{j}))^{2}}\nonumber\\
	\frac{\partial y_{r,o}}{\partial\underline{R}_{j_{w}}}&=&\frac{(1-q_{r,o})\underline{\Omega}_{j_{w},o}X_{e}(n)}{\sum_{j=1}^{K}(\underline{R}_{j}+\overline{R}_{j})}-\frac{(1-q_{r,o})\underline{R}\underline{\Omega}_{m}X_{e}(n)+q_{r,o}\overline{R}\underline{\Omega}_{m}X_{e}(n)}{(\sum_{j=1}^{K}(\underline{R}_{j}+\overline{R}_{j}))^{2}}\nonumber\\
	\frac{\partial y_{r,o}}{\partial\overline{R}_{j_{w}}}&=&\frac{q_{r,o}\overline{\Omega}_{j_{w},o}X_{e}(n)}{\sum_{j=1}^{K}(\underline{R}_{j}+\overline{R}_{j})}-\frac{(1-q_{r,o})\underline{R}\overline{\Omega}_{m}X_{e}(n)+q_{r,o}\overline{R}\overline{\Omega}_{m}X_{e}(n)}{(\sum_{j=1}^{K}(\underline{R}_{j}+\overline{R}_{j}))^{2}}\nonumber
	\end{eqnarray}
	\begin{eqnarray}
	\frac{\underline{R}_{j_w}}{\partial m_{i,j_w}}&=&\prod_{i'=1,i'\neq i}^{I}\underline{Q}_{i',j_w}(m_{i',j_w})\cdot \frac{1}{n_{s}}\sum_{r=1}^{n_s}\widetilde{\Psi}_{r}(\underline{\theta}_{i'j_{w}}^{r})\nonumber\\
	\frac{\overline{R}_{j_w}}{\partial m_{i,j_w}}&=&\prod_{i'=1,i'\neq i}^{I}\overline{Q}_{i',j_w}(m_{i',j_w})\cdot \frac{1}{n_{s}}\sum_{r=1}^{n_s}\widetilde{\Psi}_{r}(\overline{\theta}_{i'j_{w}}^{r})\nonumber\\
	\frac{\partial\underline{R}_{j_{w}}}{\underline{\theta}_{i,j_{w}}^{r}}&=&\prod_{i'=1,i'\neq i}^{I}\underline{Q}_{i',j_{w}}(\underline{\theta}_{i,j_{w}}^{r})\cdot\frac{1}{n_{s}}\widetilde{\Phi}_{r}(\underline{\theta}_{i,j_{w}}^{r})\nonumber\\
	\frac{\partial\overline{R}_{j_{w}}}{\overline{\theta}_{i,j_{w}}^{r}}&=&\prod_{i'=1,i'\neq i}^{I}\overline{Q}_{i',j_{w}}(\overline{\theta}_{i,j_{w}}^{r})\cdot\frac{1}{n_{s}}\widetilde{\Phi}_{r}(\overline{\theta}_{i,j_{w}}^{r})\nonumber
	\end{eqnarray}
	\begin{eqnarray}
	\widetilde{\Psi}_{r}(\theta)&=&\left\{ \begin{array}{cc}
	-\frac{\beta\exp(-\beta(x_{i'}-m_{i',j_{w}}+|\theta|))}{(1+\exp(-\beta(x_{i'}-m_{i'j_{w}}+|\theta|)))^{2}}, & -\infty<x_{i'}<m_{i'j_{w}}\\
	\frac{\beta\exp(-\beta(x_{i'}-m_{i',j_{w}}+|\theta|))}{(1+\exp(-\beta(x_{i'}-m_{i'j_{w}}+|\theta|)))^{2}}, & m_{i'j} \leq x_{i'}<\infty
	\end{array}\right.\nonumber \\
	\widetilde{\Phi}_{r}(\theta)&=&\left\{ \begin{array}{ccc}
	\frac{\beta\exp(-\beta(x_{i'}-m_{i',j_{w}}+\theta))}{(1+\exp(-\beta(x_{i'}-m_{i',j_{w}}+\theta)))^{2}}, & -\infty<x_{i'}<m_{i'j_{w}},\thickspace &\theta\geq 0\nonumber\\
	-\frac{\beta\exp(-\beta(x_{i'}-m_{i',j_{w}}+\theta))}{(1+\exp(-\beta(x_{i'}-m_{i',j_{w}}+\theta)))^{2}}, & m_{i'j} \leq x_{i'}<\infty,\thickspace &\theta\geq 0\nonumber\\
	-\frac{\beta\exp(-\beta(x_{i'}-m_{i',j_{w}}-\theta))}{(1+\exp(-\beta(x_{i'}-m_{i',j_{w}}-\theta)))^{2}}, & -\infty<x_{i'}<m_{i'j_{w}},\thickspace &\theta< 0\nonumber\\
	\frac{\beta\exp(-\beta(x_{i'}-m_{i',j_{w}}-\theta))}{(1+\exp(-\beta(x_{i'}-m_{i',j_{w}}-\theta)))^{2}}, & m_{i'j} \leq x_{i'}<\infty,\thickspace &\theta< 0\nonumber
	\end{array}\right.
	\end{eqnarray}
	
	\section{Experiments and Data Analysis}
	In this section, the application of eT2QFNN for RFID localization is discussed. Several experiments are conducted in the real-world environment to evaluate the efficacy of eT2QFNN embracing the MM classifier. The results are compared against 4 state-of-the-art algorithm: gClass \citep{pratama2014gclass}, pClass \citep{pratama2015pclass}, eT2Class \citep{pratama2016et2class} and eT2ELM \citep{pratama2015novel}. Five performance metrics are used, those are classification rate, the number of fuzzy rule, and the time to execute training and testing processes (execution time). The experiments are conducted under cross-validation and periodic hold-out scenario. The technical details of this experiments are elaborated in the subsection \ref{subsection:511}, while the subsection \ref{subsection:512} presents the consolidated results.
	
	\subsection{Experiment setup}
	\label{subsection:511}
\begin{figure}[t!]
		\begin{centering}
			\includegraphics[scale=0.350]{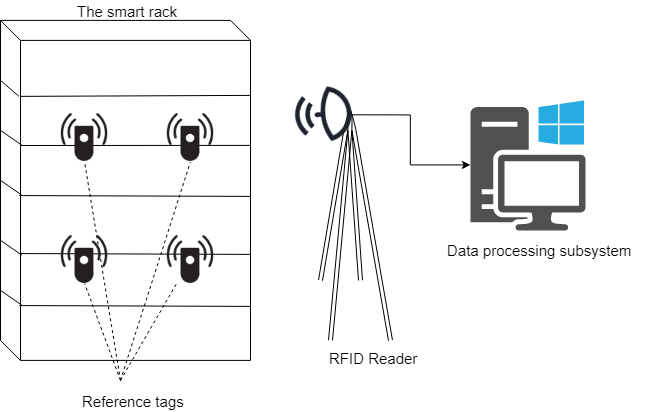}
			\par\end{centering}
		\caption{The illustration of RFID smart rack}
		\label{fig:smartrack}
	\end{figure}

	The experiments were conducted at SIMTech laboratory, Singapore. The environment is arranged to resemble the RFID smart rack system. The system utilizes RFID technology to improve the work-flow efficiency by providing the ctatic location of tools and materials for production purposes. As illustrated in the Fig. \ref{fig:smartrack}, this system consists of 1 RFID reader, 4 passive RFID tags as references which are fixed in 4 locations, and a data processing subsystem. The dimension of rack is $1510 \thickspace mm \times 600 \thickspace mm \times 2020 \thickspace mm$. The rack has 5 shelves and each shelf can load up to 6 test objects. The number of reference tag indicates that there are 4 class label considered in this experiments. The RFID reader is placed at $1000 \thickspace mm$ distance in front of the rack. The antenna is $2200\thickspace mm$ height above the ground. The reader is then connected to an RFID receiver which functions to transmit the signal into a data processing subsystem. Ethernet links is utilized to accommodate the signal transmission. Notably, one may install more reference tags and RFID reader for larger smart rack system to increase the localization accuracy \citep{chai2017reference}.
	
	The data processing subsystem has two main components, i.e. data acquisition and the algorithm execution component. The Microsoft Visual C++ based PC application is developed to acquire the RSS information data from all tags, while the localization algorithm is executed on the MATLAB 2018a online environment. The Reader is configured to report the RSS information every $1\thickspace s$. The experiment had been conducted for $20 \thickspace hours$. There are 283100 observations obtained via the experiment, each reference tag sent up to 70775 observations. It is obvious that these data obtained from the same real-time experiment, and therefore all of them pose the same distribution. Finally, the observation data can be processed to identify the object location by executing the localization algorithm.
	
	\subsection{Comparison with existing results}
	\label{subsection:512}
	To further investigate the performance of eT2QFNN, it is compared to the existing classification method, i.e. gClass, pClass, eT2Class and eT2ELM. The comparison is conducted in the same computational environments, i.e. MATLAB Online R2018a. The gClass and pClass utilize generalized type-1 fuzzy rule, while the eT2Class and eT2ELM are built upon generalized type-2 fuzzy rule. These methods are able to grow and prune its network structure according to the information provided by the current data stream. All of them except pClass is also capable to merge the similar rules. Further, the eT2ELM is encompassed with active learning and feature selection scenario which helps to discard the unnecessary training data. The eT2QFNN utilizes MM classifier, while others make use MIMO classifier. This classifier is very dependent on rule consequents because it establishes a first-order polynomial for each class. Another characteristic of this classifier is a transformation of true class label to either 0 or 1. As an illustration, if there are 4 class label and the target class is 2, then it will be converted into $[0,1,0,0]$ \citep{pratama2015pclass}.
	
	There are two experiments conducted to test the algorithm, i.e. 10-folds cross-validation and direct partition experiments. The first experiment is aimed to test the algorithms consistency while delivering the result. The experiment is started by dividing the data into 10-folds, and 9 folds data is for training, while 1 fold data is for validation. The performance metrics are achieved by averaging the results of 10-folds cross-validation. In the second experiments, the periodic hold-out evaluation scenario is conducted. The algorithms take 50000 data for training and 233100 for validation. The classification rate for the experiments is measured only in the validation phase. In contrast, the execution time is taken into account since the beginning of training phase. In this experiment, we vary $n_s=[0,10]$ and set $\eta=0.001$. The results are presented in Table \ref{tab:hasilcv} and \ref{hasil benchmark}.
	
	It can be seen from the Table \ref{tab:hasilcv} that the eT2QFNN delivers most reliable classification rates. Although it employs 4 sub-models to obtain this result which burden the computation, eT2QFNN has the fastest execution time second only to pClass. In terms of network complexity, eT2QFNN generates a comparable number of the fuzzy rule. It worth noting that eT2QFNN is not encompassed with the rule merging and pruning scenario. Further, the number of eT2QFNN rule is less than eT2ELM which has rule pruning and merging scenario. Table \ref{hasil benchmark} confirms the consistency of eT2QFNN while delivering good result. It worth noting that the second experiment utilizes less data for training, however eT2QFNN maintains the classification rate around $97\%$ which is still comparable to another methods. The execution time is lower than others method except eT2ELM. It is obvious because eT2ELM has the online active scenario which can reduce the training sample.
	\begin{table}
		\caption{The results of the cross-validation experiment compare to the benchmarked
			algorithms} \label{tab:hasilcv}
		\centering{}%
		\begin{tabular}{p{3cm}p{3cm}p{2.5cm}}
			\hline\noalign{\smallskip}
			Algorithms & \multicolumn{2}{c}{Results}\tabularnewline
			\noalign{\smallskip}\svhline\noalign{\smallskip}
			\multirow{3}{*}{MM-eT2QFNN} & Classification rate & \textbf{0.99}$\thickspace \pm \thickspace$\textbf{0.05}\tabularnewline
			& Rule & 6.23$\thickspace \pm \thickspace$0.68\tabularnewline
			& Execution time & \textbf{618.63}$\thickspace \pm \thickspace$\textbf{31.64} \tabularnewline
			\noalign{\smallskip}\hline\noalign{\smallskip}
			\multirow{3}{*}{gClass} & Classification rate & 0.97 $\thickspace \pm \thickspace$ 0.006 \tabularnewline
			& Rule & 2.4 $\thickspace \pm \thickspace$ 1.2 \tabularnewline
			& Execution time & 1004.36 $\thickspace \pm \thickspace$ 97.78 \tabularnewline
			\noalign{\smallskip}\hline\noalign{\smallskip}
			\multirow{3}{*}{pClass} & Classification rate & 0.97 \tabularnewline
			& Rule & 2 \tabularnewline
			& Execution time & \textbf{369.28} $\thickspace \pm \thickspace$ \textbf{9.99} \tabularnewline
			\noalign{\smallskip}\hline\noalign{\smallskip} 
			\multirow{3}{*}{eT2Class} & Classification rate & 0.97 $\thickspace \pm \thickspace$ 0.008 \tabularnewline
			& Rule & 2 \tabularnewline
			& Execution time & 447.36 $\thickspace \pm \thickspace$ 11.60 \tabularnewline
			\noalign{\smallskip}\hline\noalign{\smallskip} 
			\multirow{3}{*}{eT2ELM} & Classification rate & 0.95 $\thickspace \pm \thickspace$ 0.018 \tabularnewline
			& Rule & 37.2 $\thickspace \pm \thickspace$ 5.95 \tabularnewline
			& Execution time & 1324.1 $\thickspace \pm \thickspace$ 109.47  \tabularnewline
			\noalign{\smallskip}\hline\noalign{\smallskip} 
		\end{tabular}
	\end{table}
	\begin{table}
		\caption{The results of the direct partition experiment compare to the benchmarked
			algorithms}\label{hasil benchmark}
		\centering{}%
		\begin{tabular}{p{3cm}p{3cm}p{1cm}}
			\hline\noalign{\smallskip}
			Algorithms & \multicolumn{2}{c}{Results}\tabularnewline
			\noalign{\smallskip}\svhline\noalign{\smallskip}
			\multirow{3}{*}{MM-eT2QFNN} & Classification rate & 0.97 \tabularnewline
			& Rule & \textbf{4.75} \tabularnewline
			& Execution time & \textbf{131.5}\tabularnewline
			\noalign{\smallskip}\hline\noalign{\smallskip}
			\multirow{3}{*}{gClass} & Classification rate & 0.99\tabularnewline
			& Rule & 4\tabularnewline
			& Execution time & 290.83\tabularnewline
			\noalign{\smallskip}\hline\noalign{\smallskip} 
			\multirow{3}{*}{pClass} & Classification rate & 0.98\tabularnewline
			& Rule & 2\tabularnewline
			& Execution time & 225.8\tabularnewline
			\noalign{\smallskip}\hline\noalign{\smallskip}
			\multirow{3}{*}{eT2Class} & Classification rate & 0.98\tabularnewline
			& Rule & 2\tabularnewline
			& Execution time & 330.71\tabularnewline
			\noalign{\smallskip}\hline\noalign{\smallskip} 
			\multirow{3}{*}{eT2ELM} & Classification rate & 0.98\tabularnewline
			& Rule & \textbf{5}\tabularnewline
			& Execution time & \textbf{41.73} \tabularnewline
			\noalign{\smallskip}\hline\noalign{\smallskip} 
		\end{tabular}
	\end{table}
	\section{Conclusions}
	This paper presents an evolving model based on the EIS, namely eT2QFNN. The fuzzification layer relies on IT2QMF, which has a graded membership degree and footprints of uncertainties. The IT2QMF is the extended version of QMF which are able to both capture the input uncertainties and to identify overlaps between input classes. The eT2QFNN works fully in the evolving mode, that is the network parameters and the number of rules are adjusted and generated on the fly. The parameter adjustment scenario is achieved via DEKF. Meanwhile, the rule growing mechanism is conducted by measuring the statistical contribution of the hypothetical rule. The new rule is formed when its statistical contribution is higher than the sum of others multiplied by vigilance parameter. The proposed method is utilized to predict the class label of an object according to the RSS information provided by the reference tags. The conducted experiments simulates the RFID smart rack system which is constructed by 4 reference tags, 1 RFID reader and a data processing subsystem to execute the algorithm. The experiments results show that eT2QFNN is capable of delivering comparable accuracy benchmarked to state-of-the-art algorithms while maintaining low execution time and compact network.
	\begin{acknowledgement}
		This project is fully supported by NTU start-up grant and Ministry of Education Tier 1 Research Grant. We also would like to thank Singapore Institute of Manufacturing Technology which provided the RFID data that greatly assisted the research.
	\end{acknowledgement}
	%\section*{Appendix}
	%\addcontentsline{toc}{section}{Appendix}
	%When placed at the end of a chapter or contribution (as opposed to at the end of the book), the numbering of tables, figures, and equations in the appendix section continues on from that in the main text. Hence please \textit{do not} use the \verb|appendix| command when writing an appendix at the end of your chapter or contribution. If there is only one the appendix is designated ``Appendix'', or ``Appendix 1'', or ``Appendix 2'', etc. if there is more than one.
	
	\bibliographystyle{spbasic}
	\bibliography{reference}
\end{document}